\documentclass[runningheads]{llncs}
\usepackage{graphicx}
\usepackage{amsmath,amssymb} 

\usepackage{color}
\usepackage{mathtools}
\usepackage{epsfig}
\DeclarePairedDelimiter\floor{\lfloor}{\rfloor}

\usepackage[width=122mm,left=12mm,paperwidth=146mm,height=193mm,top=12mm,paperheight=217mm]{geometry}
\begin{document}
\pagestyle{headings}
\mainmatter
\def\ECCV18SubNumber{1817}  

\title{ExFuse: Enhancing Feature Fusion for Semantic Segmentation} 

\titlerunning{ExFuse: Enhancing Feature Fusion for Semantic Segmentation}

\authorrunning{Zhang {\em et\ al.}}


\author{Zhenli Zhang$^1$, Xiangyu Zhang$^2$, Chao Peng$^2$, Dazhi Cheng$^3$, Jian Sun$^2$
}

\institute{ ${}^1$Fudan University, ${}^2$Megvii Inc. (Face++), ${}^3$Beijing Institute of Technology
{\tt\small zhenlizhang14@fudan.edu.cn}, \{{\tt\small zhangxiangyu, pengchao}\}{\tt\small@megvii.com}, {\tt\small1120152020@bit.edu.com}, {\tt\small sunjian@megvii.com}
}

\maketitle

\begin{abstract}
Modern semantic segmentation frameworks usually combine low-level and high-level features from pre-trained backbone convolutional models to boost performance. In this paper, we first point out that a simple fusion of low-level and high-level features could be less effective because of the gap in semantic levels and spatial resolution. We find that introducing semantic information into low-level features and high-resolution details into high-level features is more effective for the later fusion. Based on this observation, we propose a new framework, named ExFuse, to bridge the gap between low-level and high-level features thus significantly improve the segmentation quality by 4.0\% in total. Furthermore, we evaluate our approach on the challenging PASCAL VOC 2012 segmentation benchmark and achieve 87.9\% mean IoU, which outperforms the previous state-of-the-art results.
\end{abstract}

\section{Introduction}
\label{sec:introduction}

\begin{figure}[htbp]
	\centering
	\includegraphics[width=0.60\linewidth]{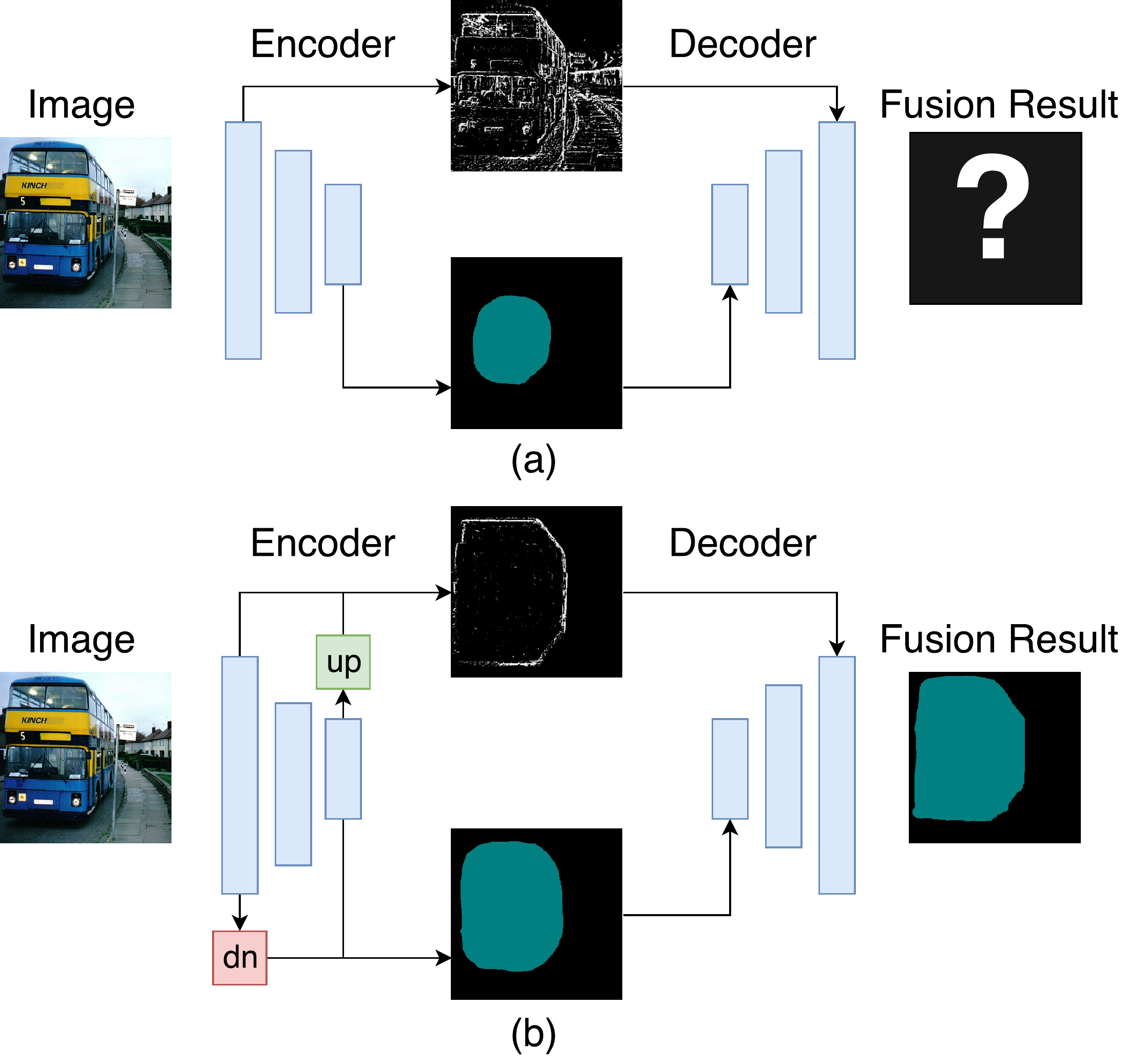}
	\caption{Fusion of low-level and high-level features. a) ``Pure'' low-level high-resolution and ``pure'' high-level low-resolution features are difficult to be fused because of the significant semantic and resolution gaps. b) Introducing semantic information into low-level features or spatial information into high-level features benefits the feature fusion. ``dn'' and ``up'' blocks represent \emph{abstract} up/down-sampling feature embedding.}
	\label{fig:concept}
\end{figure}

Most state-of-the-art semantic segmentation frameworks \cite{Chen2016DeepLab,Ghiasi2016Laplacian,Badrinarayanan2017SegNet,Yu2015Multi,Chen2017Rethinking,Islam_2017_CVPR,Ronneberger2015U,Peng2017Large,Wang2017Understanding,Zhao2016Pyramid,Lin2016RefineNet,Chen2014Semantic} follow the design of Fully Convolutional Network (FCN) \cite{Long2015Fully}. FCN has a typical encoder-decoder structure -- semantic information is firstly embedded into the feature maps via encoder then the decoder takes responsibility for generating segmentation results. Usually the encoder is the pre-trained convolutional model to extract image features and the decoder contains multiple upsampling components to recover resolution. Although the top-most feature maps of the encoder could be highly semantic, its ability to reconstruct precise details in segmentation maps is limited due to insufficient resolution, which is very common in modern backbone models such as \cite{He2016Deep,Szegedy2015Going,Krizhevsky2012ImageNet,He2016Identity,Xie2016Aggregated,Simonyan2014Very}. To address this, an ``U-Net'' architecture is proposed \cite{Ronneberger2015U} and adopted in many recent work \cite{Ghiasi2016Laplacian,Islam_2017_CVPR,Ronneberger2015U,Peng2017Large,Long2015Fully,Lin2016RefineNet}. The core idea of \emph{U-Net} is to gradually fuse high-level low-resolution features from top layers with low-level but high-resolution features from bottom layers, which is expected to be helpful for the decoder to generate high-resolution semantic results.

Though the great success of U-Net, the working mechanism is still unknown and worth further investigating. Low-level and high-level features are complementary by nature, where low-level features are rich in spatial details but lack semantic information and vice versa. Consider the extreme case that ``pure'' low-level features only encode low-level concepts such as points, lines or edges. Intuitively, the fusion of high-level features with such ``pure'' low-level features helps little, because low-level features are too noisy to provide sufficient high-resolution semantic guidance. In contrast, if low-level features include more semantic information, for example, encode relatively clearer semantic boundaries, then the fusion becomes easy -- fine segmentation results could be obtained by aligning high-level feature maps to the boundary. Similarly, ``pure'' high-level features with little spatial information cannot take full advantage of low-level features; however, with additional high-resolution features embedded, high-level features may have chance to refine itself by aligning to the nearest low-level boundary. Fig~\ref{fig:concept} illustrates the above concepts. Empirically, the semantic and resolution overlap between low-level and high-level features plays an important role in the effectiveness of feature fusion. In other words, feature fusion could be enhanced by introducing more semantic concepts into low-level features or by embedding more spatial information into high-level features.

Motivated by the above observation, we propose to boost the feature fusion by bridging the semantic and resolution gap between low-level and high-level feature maps. We propose a framework named \emph{ExFuse}, which addresses the gap from the following two aspects: 1) to introduce more semantic information into low-level features, we suggest three solutions -- \emph{layer rearrangement}, \emph{semantic supervision} and \emph{semantic embedding branch}; 2) to embed more spatial information into high-level features, we propose two novel methods: \emph{explicit channel resolution embedding} and \emph{densely adjacent prediction}. Significant improvements are obtained by either approach and a total increase of 4\% is obtained by the combination. Furthermore, we evaluate our method on the challenging PASCAL VOC 2012 \cite{Everingham2010The} semantic segmentation task. In the test dataset, we achieve the score of 87.9\% mean IoU, surpassing the previous state-of-the-art methods.

Our contributions can be summerized as follows:
\begin{itemize}
\item We suggest a new perspective to boost semantic segmentation performance, i.e. bridging the semantic and resolution gap between low-level and high-level features by more effective feature fusion.
\item We propose a novel framework named ExFuse, which introduces more semantic information into low-level features and more spatial high-resolution information into high-level features. Significant improvements are obtained from the enhanced feature fusion.
\item Our fully-equipped model achieves the new state-of-the-art result on the test set of PASCAL VOC 2012 segmentation benchmark. 
\end{itemize}

\section{Related Work}

\paragraph{Feature fusion in semantic segmentation.}
Feature fusion is frequently employed in semantic segmentation for different purposes and concepts. A lot of methods fuse low-level but high-resolution features and high-level low-resolution features together \cite{Ghiasi2016Laplacian,Islam_2017_CVPR,Ronneberger2015U,Peng2017Large,Long2015Fully,Lin2016RefineNet}. Besides, \emph{ASPP} module is proposed in DeepLab \cite{Chen2016DeepLab,Chen2017Rethinking,Chen2014Semantic} to fuse multi-scale features to tackle objects of different size. Pyramid pooling module in PSPNet \cite{Zhao2016Pyramid} serves the same purpose through different implementation. BoxSup \cite{Dai2015BoxSup} empirically fuses feature maps of bounding boxes and segmentation maps to further enhance segmentation.

\paragraph{Deeply supervised learning.}
To the best of our knowledge, deeply supervised training is initially proposed in \cite{Lee2014Deeply}, which aims to ease the training process of very deep neural networks since depth is the key limitation for training modern neural networks until batch normalization \cite{Ioffe2015Batch} and residual networks \cite{He2016Deep} are proposed. Extra losses are utilized in GoogleNet \cite{Szegedy2015Going} for the same purpose. Recently, PSPNet \cite{Zhao2016Pyramid} also employs this method to ease the optimization when training deeper networks.

\paragraph{Upsampling.}
There are mainly three approaches to upsample a feature map. The first one is bilinear interpolation, which is widely used in \cite{Chen2016DeepLab,Chen2017Rethinking,Zhao2016Pyramid,Chen2014Semantic}. The second method is deconvolution, which is initially proposed in FCN \cite{Long2015Fully} and utilized in later work such as \cite{Badrinarayanan2017SegNet,Islam_2017_CVPR,Ronneberger2015U,Peng2017Large,Lin2016RefineNet}. The third one is called ``sub-pixel convolution", which derives from \cite{Shi2016Real,Aitken2017Checkerboard} in super resolution task and is widely broadcast to other tasks such as semantic segmentation. For instance, \cite{Wang2017Understanding} employs it to replace the traditional deconvolution operation.

\section{Approach}
\label{sec:approach}

\begin{figure*}[t]
\begin{center}
\includegraphics[width=1\linewidth]{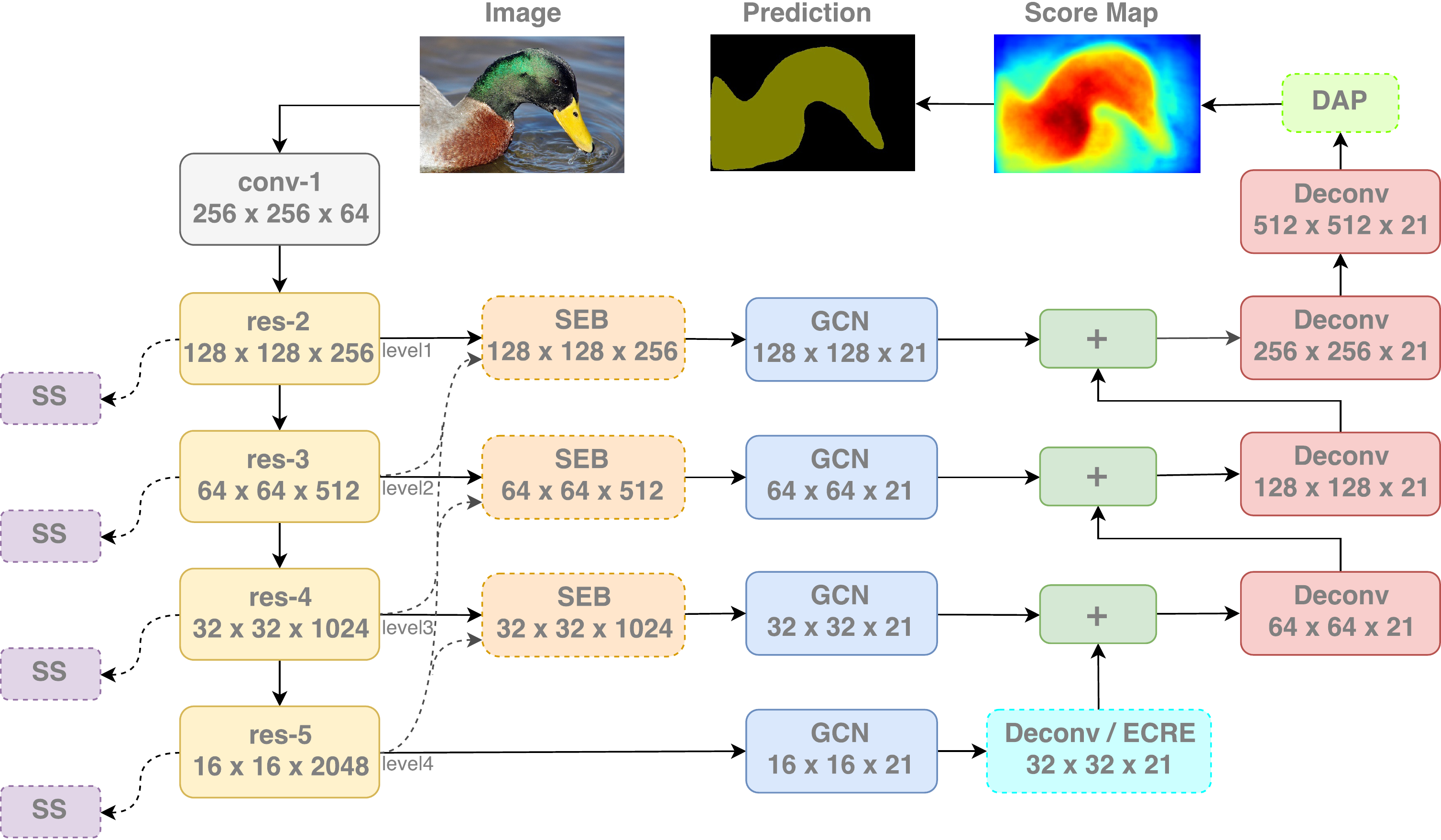}
\end{center}
   \caption{Overall architecture of our approach. Components with solid boxes belong to the backbone \emph{GCN} framework \cite{Peng2017Large}, while others with dashed lines are proposed in this work. Similar to \cite{Peng2017Large}, \emph{Boundary Refinement} blocks are actually used but omitted in the figure. Numbers ($H\times W \times C$) in blocks specify the output dimension of each component. \textbf{SS} -- semantic supervision. \textbf{ECRE} -- explicit channel resolution embedding. \textbf{SEB} -- semantic embedding branch. \textbf{DAP} -- densely adjacent prediction.}
\label{fig:arch}
\end{figure*}

In this work we mainly focus on the feature fusion problem in ``U-Net'' segmentation frameworks \cite{Ghiasi2016Laplacian,Islam_2017_CVPR,Ronneberger2015U,Peng2017Large,Long2015Fully,Lin2016RefineNet}. In general, \emph{U-Net} have an encoder-decoder structure as shown in Fig~\ref{fig:concept}. Usually the encoder part is based on a convolutional model pretrained on large-scale classification dataset (e.g. ImageNet \cite{Deng2009ImageNet}), which generates low-level but high-resolution features from the bottom layers and high-level low-resolution features from the top layers. Then the decoder part mixes up the features to predict segmentation results. A common way of feature fusion \cite{Pohlen2016Full,Ghiasi2016Laplacian,Hariharan2014Hypercolumns,Islam_2017_CVPR,Ronneberger2015U,Peng2017Large,Lin2016RefineNet} is to formulate as a residual form:
\begin{equation}
\mathbf{y}_l=Upsample(\mathbf{y}_{l+1})+\mathcal{F}(\mathbf{x}_l)
\label{equ:basicfuse}
\end{equation}
where $\mathbf{y}_l$ is the fused feature at $l$-th level; $\mathbf{x}_l$ stands for the $l$-th feature generated by the encoder. Features with larger $l$ have higher semantic level but lower spatial resolution and vice versa (see Fig~\ref{fig:arch}). 

In Sec \ref{sec:introduction} we argue that feature fusion could become less effective if there is a large semantic or resolution gap between low-level and high-level features. To study and verify the impact, we choose one of the start-of-the-art ``U-Net'' frameworks -- \emph{Global Convolutional Network (GCN)} \cite{Peng2017Large} -- as our backbone segmentation architecture (see Fig~\ref{fig:arch} for details). In GCN, 4 different semantic levels of feature maps are extracted from the encoder network, whose spatial resolutions, given the $512\times 512$ input, are $\{128,64,32,16\}$ respectively. To examine the effectiveness of feature fusion, we select several subsets of feature levels and use them to retrain the whole system. Results are shown in Table \ref{tbl:basicfusion}. It is clear that even though the segmentation quality increases with the fusion of more feature levels, the performance tends to saturate quickly. Especially, the lowest two feature levels (1 and 2) only contribute marginal improvements (0.24\% for ResNet 50 and 0.05\% for ResNeXt 101), which implies the fusion of low-level and high-level features is rather ineffective in this framework.

In the following subsections we will introduce our solutions to bridge the gap between low-level and high-level features -- embedding more semantic information into low-level features and more spatial resolution clues into high-level features. First of all, we introduce our baseline settings:
\begin{table}

\begin{center}
\begin{tabular}{c|c|c}
\hline
Feature Levels & ResNet 50 (\%) & ResNeXt 101 (\%) \\
\hline
$\{4\}$ & 70.04 & 73.79 \\
\hline
$\{3,4\}$ & 72.17& 75.97\\
\hline
$\{2,3,4\}$ & 72.28& 75.98\\
\hline
$\{1,2,3,4\}$ & 72.41& 76.02\\
\hline
\end{tabular}
\end{center}

\caption{\emph{GCN} \cite{Peng2017Large} segmentation results using given feature levels. Performances are evaluated by standard mean IoU(\%) on PASCAL VOC 2012 validation set. Lower feature level involves less semantic but higher-resolution features and vice versa (see Fig~\ref{fig:arch}). The feature extractor is based on pretrained ResNet50 \cite{He2016Deep} and ResNeXt101\cite{Xie2016Aggregated} model. Performance is evaluated in mIoU.}

\label{tbl:basicfusion}
\end{table}

\paragraph{Baseline Settings.} 
The overall semantic segmentation framework follows the fully-equipped \emph{GCN} \cite{Peng2017Large} architecture, as shown in Fig~\ref{fig:arch}. For the backbone encoder network, we use ResNeXt 101 \cite{Xie2016Aggregated} model pretrained on ImageNet by default\footnote{Though ResNeXt 101 performs much better than ResNet 101\cite{He2016Deep} on ImageNet classification task (21.2\% vs. 23.6\% in top-1 error), we find there are no significant differences on the semantic segmentation results (both are 76.0\% mIoU). } unless otherwise mentioned. We use two public-available semantic segmentation benchmarks -- \emph{PASCAL VOC 2012} \cite{Everingham2010The} and \emph{Semantic Boundaries Dataset} \cite{Hariharan2011Semantic} -- for training and evaluate performances on PASCAL VOC 2012 validation set, which is consistent with many previous work \cite{Chen2016DeepLab,Ghiasi2016Laplacian,Badrinarayanan2017SegNet,Yu2015Multi,Chen2017Rethinking,Islam_2017_CVPR,Peng2017Large,Wang2017Understanding,Zhao2016Pyramid,Long2015Fully,Lin2016RefineNet,Chen2014Semantic,Pohlen2016Full}. The performance is measured by standard mean intersection-over-union (mean IoU). Other training and test details or hyper-parameters are exactly the same as \cite{Peng2017Large}. Our reproduced GCN baseline score is 76.0\%, shown in Table \ref{tbl:ablation} (\#1).

\subsection{Introducing More Semantic Information into Low-level Features}
\label{sec:tolowlevel}

Our solutions are inspired by the fact: for convolutional neural networks, feature maps close to semantic supervisions (e.g. classification loss) tend to encode more semantic information, which has been confirmed by some visualization work \cite{zeiler2014visualizing}. We propose three methods as follows:

\subsubsection{Layer Rearrangement}
\label{sec:lr}

In our framework, features are extracted from the tail of each stage in the encoder part (res-2 to res-5 in Fig~\ref{fig:arch}). To make low-level features (res-2 or res-3) 'closer' to the supervisions, one straight-forward approach is to arrange more layers in the early stages rather than the latter. For example, ResNeXt 101 \cite{Xie2016Aggregated} model has $\{3,4,23,3\}$ building blocks for Stage 2-5 respectively; we rearrange the assignment into $\{8,8,9,8\}$ and adjust the number of channels to ensure the same overall computational complexity. Experiment shows that even though the ImageNet classification score of the newly designed model is almost unchanged, its segmentation performance increases by 0.8\% (Table~\ref{tbl:ablation}, compare \#2 with \#3), which implies the quality of low-level feature might be improved.

\subsubsection{Semantic Supervision}
\label{sec:semanticsupervision}

We come up with another way to improve low-level features, named \emph{Semantic Supervision (SS)}, by assigning auxiliary supervisions directly to the early stages of the encoder network (see Fig~\ref{fig:arch}). To generate semantic outputs in the auxiliary branches, low-level features are forced to encode more semantic concepts, which is expected to be helpful for later feature fusion. Such methodology is inspired by \emph{Deeply Supervised Learning} used in some old classification networks \cite{Lee2014Deeply,Szegedy2015Going} to ease the training of deep networks. However, more sophisticated classification models \cite{szegedy2016rethinking,szegedy2017inception,He2016Deep,He2016Identity,hu2017squeeze,Xie2016Aggregated} suggest end-to-end training without auxiliary losses, which is proved to have no convergence issue even for models over 100 layers. Our experiment also shows that for ResNet or ResNeXt models deeply supervised training is useless or even harms the classification accuracy (see Table~\ref{tbl:semanticsup}). Therefore, our \emph{Semantic Supervision} approach mainly focuses on improving the quality of low-level features, rather than boosting the backbone model itself.

\begin{table}
\begin{center}
\begin{tabular}{l|c|c}
\hline
Model & Cls err (top-1, \%) & Seg mIoU (\%) \\
\hline
Res50 & 24.15 & 72.4 \\
\hline
\emph{SS} Res50 & 24.77 & 73.5\\
\hline
\end{tabular}
\end{center}

\caption{Effects of \emph{Semantic Supervision (SS)}. Classification scores are evaluated on ImageNet 2012 validation set.}

\label{tbl:semanticsup}
\end{table}

\begin{figure}
	\centering
	\includegraphics[width=0.3\linewidth]{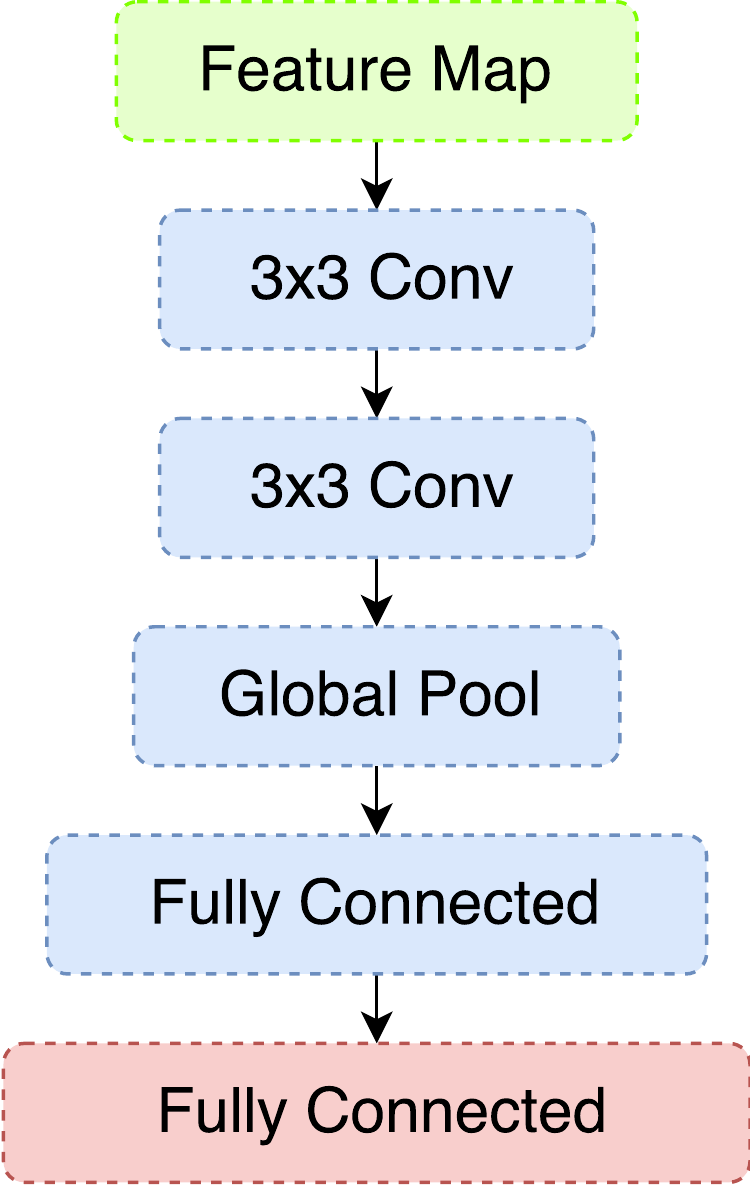}
	\caption{Details of \emph{Semantic Supervision (SS)} component in our pipeline.}
	\label{fig:semanticsupervision}
\end{figure}

Fig~\ref{fig:semanticsupervision} shows the detailed structure of our Semantic Supervision block. When pretraining the backbone encoder network, the components are attached to the tail of each stage as auxiliary supervisions (see Fig~\ref{fig:arch}). The overall classification loss equals to a weighted summation of all auxiliary branches. Then after pretraining, we remove these branches and use the remaining part for fine tuning. Experiment shows the method boosts the segmentation result by 1.1\%. Moreover, we find that if features are extracted from the second convolutional layer in the auxiliary module for fine tuning (Fig~\ref{fig:semanticsupervision}), more improvement (1.5\%) is obtained (see Table \ref{tbl:ablation}, compare \#1 with \#2), which supports our intuition that feature maps closer to the supervision tend to encode more semantic information.

It is worth noting that the recent semantic segmentation work \emph{PSPNet} \cite{Zhao2016Pyramid} also employs deeply supervised learning and reports the improvements. Different from ours, the architecture of \cite{Zhao2016Pyramid} do not extract feature maps supervised by the auxiliary explicitly; and their main purpose is to ease the optimization during training. However, in our framework we find the improvements may result from different reasons. For instance, we choose a relatively shallower network ResNet 50 \cite{He2016Deep} and pretrain with or without semantic supervision. From Table \ref{tbl:semanticsup}, we find the auxiliary losses do not improve the classification score, which implies ResNet 50 is unlikely to suffer from optimization difficulty. However, it still boosts the segmentation result by 1.1\%, which is comparable to the deeper case of ResNeXt 101 (1.0\%). We believe the enhancement in our framework mainly results from more ``semantic'' low-level features. 

\subsubsection{Semantic Embedding Branch}

As mentioned above, many ``U-Net'' structures involve low-level feature as the residue to the upsampled high-level feature. In Equ \ref{equ:basicfuse} the residual term $\mathcal{F}(\mathbf{x}_l)$ is a function of low-level but high-resolution feature, which is used to fill the spatial details. However, if the low-level feature contains little semantic information, it is insufficient to recover the semantic resolution. To address the drawback, we generalize the fusion as follows:
\begin{equation}
\mathbf{y}_l=Upsample\left(\mathbf{y}_{l+1})+\mathcal{F}(\mathbf{x}_l,\mathbf{x}_{l+1}, \ldots, \mathbf{x}_{L}\right)
\label{equ:semanticfuse}
\end{equation}
where $L$ is the number of feature levels. Our insight is to involve more semantic information from high-level features to guide the resolution fusion. 

\begin{figure}
	\centering
	\includegraphics[width=0.4\linewidth]{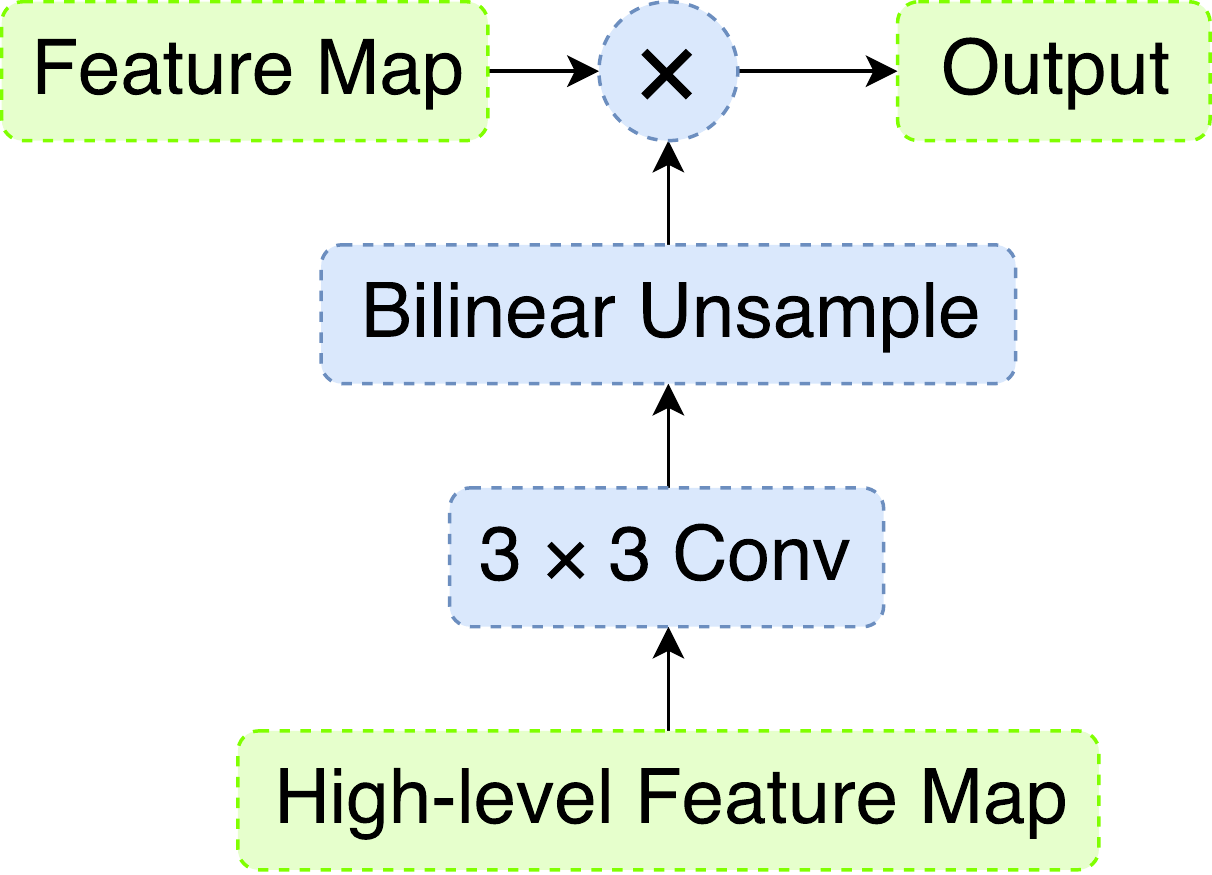}
	\caption{Design of the \emph{Semantic Embedding Branch} in Fig~\ref{fig:arch}. The ``$\times$'' sign means element-wise multiplication. If there are more than one groups of high-level features, the component outputs the production of each feature map after upsampling. }
	\label{fig:semanticembeddingbranch}
\end{figure}

The detailed design of function $\mathcal{F}\left(\cdot\right)$ is illustrated in Fig~\ref{fig:semanticembeddingbranch}, named \emph{Semantic Embedding Branch, (SEB)}. We use the component for features of Level 1-3 (see Fig~\ref{fig:arch}). In our experiment SEB improves the performance by 0.7\% (Table~\ref{tbl:ablation}, compare \#3 with \#5).

\subsection{Embedding More Spatial Resolution into High-level Features}
\label{sec:tohighlevel}

For most backbone feature extractor networks, high-level features have very limited spatial resolution. For example, the spatial size of top-most feature map in ResNet or ResNeXt is $7\times 7$ for $224\times 224$ input size. To encode more spatial details, a widely used approach is \emph{dilated strategy} \cite{Yu2015Multi,Chen2016DeepLab,Chen2017Rethinking,Wang2017Understanding,Zhao2016Pyramid,Chen2014Semantic}, which is able to enlarge feature resolution without retraining the backbone network. However, since high-level feature maps involve a lot of channels, larger spatial size significantly increases the computational cost. So in this work we mainly consider another direction -- we do not try to increase the ``physical'' resolution of the feature maps; instead, \textbf{we expect more resolution information encoded within channels}. We propose the following two methods:

\begin{table}[b]
\begin{center}
\begin{tabular}{c|cccccc|c}
\hline
Index & Baseline & SS & LR & ECRE & SEB & DAP & mIoU (\%) \\
\hline
1 & \checkmark & & & & & & 76.0 \\
\hline
2 & \checkmark & \checkmark & & & & & 77.5 \\
\hline
3 & \checkmark & \checkmark & \checkmark & & & & 78.3 \\
\hline
4 & \checkmark & \checkmark & \checkmark & \checkmark & & & 78.8 \\
\hline
5 & \checkmark & \checkmark & \checkmark & & \checkmark & & 79.0 \\
\hline
6 & \checkmark & \checkmark & \checkmark & & \checkmark & \checkmark & 79.6 \\
\hline
7 & \checkmark & \checkmark & \checkmark & \checkmark & \checkmark & \checkmark & \textbf{80.0} \\
\hline
\end{tabular}
\end{center}

\caption{Ablation experiments of the methods in Sec~\ref{sec:approach}. Performances are evaluated by standard mean IoU(\%) on PASCAL VOC 2012 validation set. The baseline model is \cite{Peng2017Large} (our impl.) \textbf{SS} -- semantic supervision. \textbf{LR} -- layer rearrangement. \textbf{ECRE} -- explicit channel resolution embedding. \textbf{SEB} -- semantic embedding branch. \textbf{DAP} -- densely adjacent prediction. }

\label{tbl:ablation}
\end{table}

\subsubsection{Explicit Channel Resolution Embedding}

In our overall framework, segmentation loss is only connected to the output of decoder network (see Fig~\ref{fig:arch}), which is considered to have less impact on the spatial information of high-level features by intuition. One straight-forward solution is to borrow the idea of \emph{Semantic Supervision} (Sec \ref{sec:semanticsupervision}) -- we could add an auxiliary supervision branch to the high-level feature map, upsample and force it to learn fine segmentation map. Following the insight, firstly we try adding an extra segmentation loss to the first deconvolution module (the light-blue component in Fig~\ref{fig:arch}), however, no improvements are obtained (Table~\ref{tbl:ecre}, \#2). 

\begin{figure}[h]
	\centering
	\includegraphics[width=0.9\linewidth]{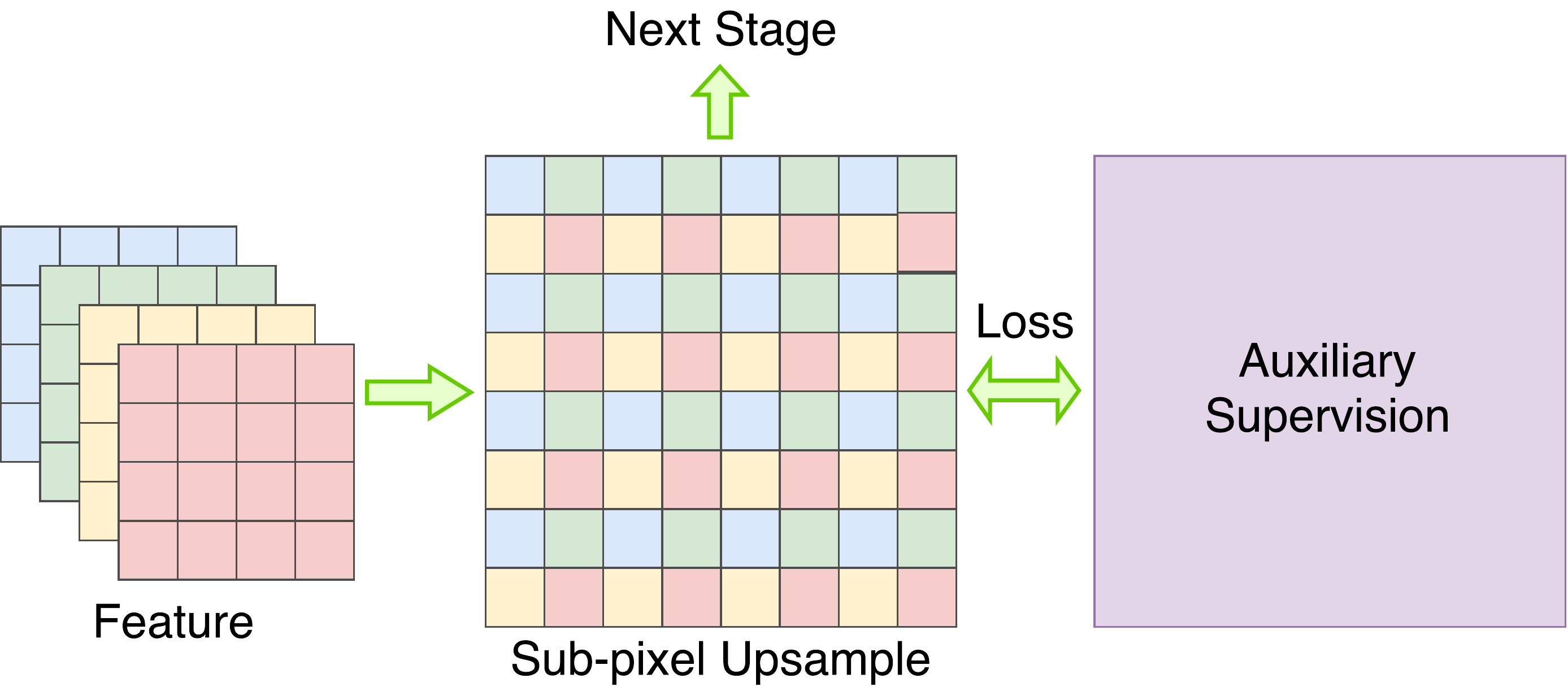}
	\caption{Illustration of the design of \emph{Explicit Channel Resolution Embedding (ECRE)} module in Fig~\ref{fig:arch}.}
	\label{fig:ecre}
\end{figure}

Why does the auxiliary loss fail to work? Note that the purpose of the supervision is to embed high resolution information ``explicitly'' into feature map channels. However, since deconvolution layer includes weights, the embedding becomes implicit. To overcome this issue, we adopt a parameter-free upsampling method -- \emph{Sub-pixel Upsample} \cite{Shi2016Real,Aitken2017Checkerboard} -- to replace the original deconvolution. Since sub-pixel upsample enlarge the feature map just by reshaping the spatial and channel dimensions, the auxiliary supervision is able to \emph{explicitly} impact the features. Details of the component are shown in Fig~\ref{fig:ecre}. Experiment shows that it enhances the performance by 0.5\% (see Table~\ref{tbl:ecre} and Table~\ref{tbl:ablation}).

\begin{table}[h]

\begin{center}
\begin{tabular}{c|c|c}
\hline
Index & Method  & mIoU (\%) \\
\hline
1 & Baseline & 78.3 \\
\hline
2 & Deconv + Supervised & 78.2\\
\hline
3 & Sub-pixel Upsample Only  & 77.6\\
\hline
4 & ECRE (Fig~\ref{fig:ecre}) & \textbf{78.8}\\
\hline
\end{tabular}
\end{center}

\caption{Ablation study on the design of \emph{Explicit Channel Resolution Embedding, (ECRE)}. The baseline model is in Table~\ref{tbl:ablation} (\#3)}

\label{tbl:ecre}
\end{table}

Moreover, to demonstrate that the improvement is brought by explicit resolution embedding rather than sub-pixel upsampling itself, we also try to replace the deconvolution layer only without auxiliary supervision. Table~\ref{tbl:ecre} (\#3) shows the result, which is even worse than the baseline.

\subsubsection{Densely Adjacent Prediction}

\begin{figure}[htbp]
	\centering
	\includegraphics[width=0.6\linewidth]{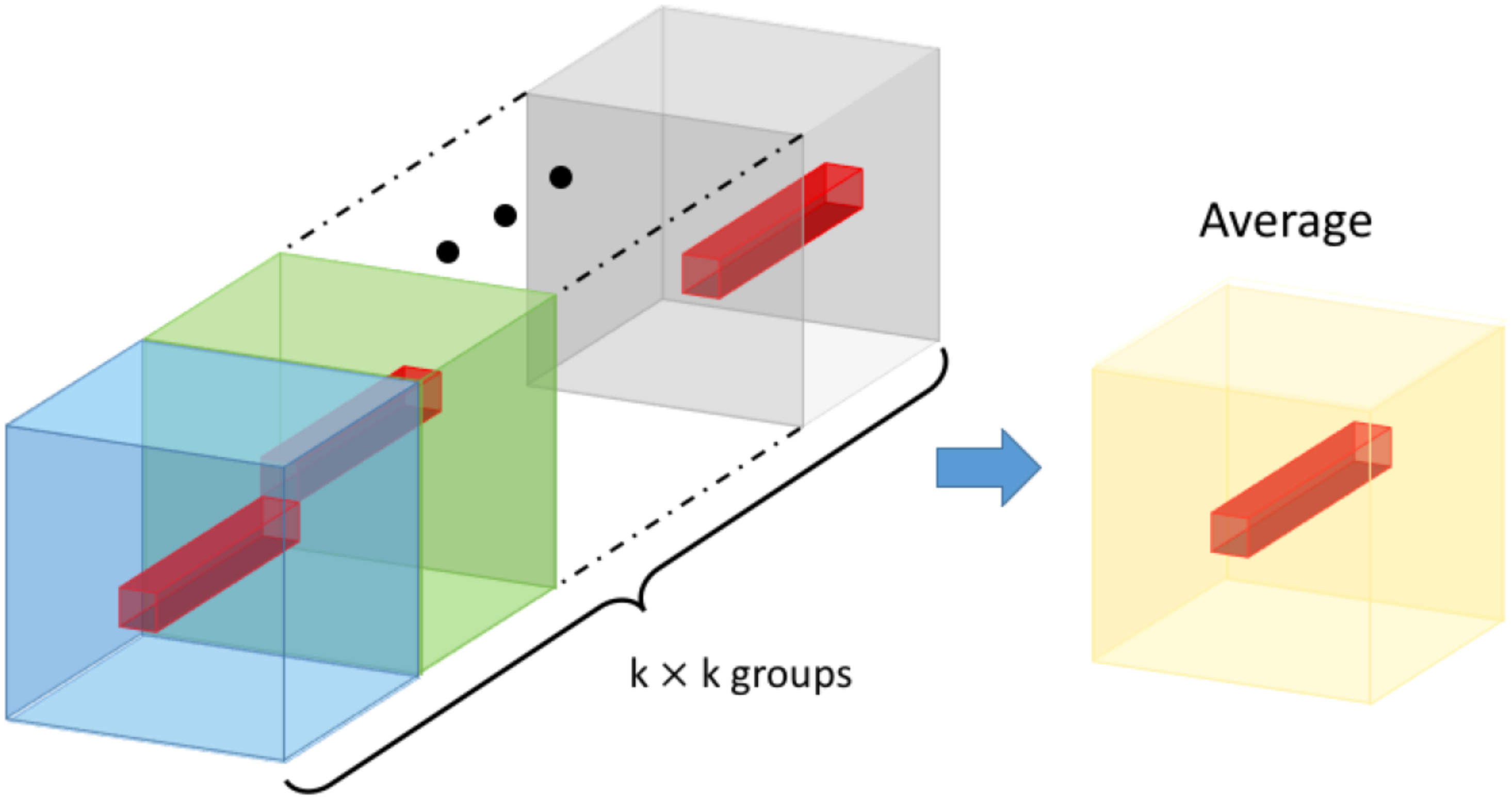}
	\caption{Illustration of \emph{Densely Adjacent Prediction (DAP)} component in Fig~\ref{fig:arch}.}
	\label{fig:dap}
\end{figure}

In the decoder upstream of the original architecture (Fig~\ref{fig:arch}), feature point at the spatial location $(i, j)$ mainly takes responsibility for the semantic information at the same place. To encode as much spatial information into channels, we propose a novel mechanism named \emph{Densely Adjacent Prediction (DAP)}, which allows to predict results at the adjacent position, e.g. $(i-1, j+1)$. Then to get the final segmentation map, result at the position $(i, j)$ can be generated by averaging the associated scores. Formally, given the window size $k\times k$, we divide the feature channels into $k\times k$ groups, then {DAP} works as follows:
\begin{equation}
\mathbf{r}_{i,j} = \frac{1}{k\times k}\sum _{0\leq l,m<k} \mathbf{x}_{i+l-\floor*{k/2}, j+m-\floor*{k/2}}^{(l\times k+m)}
\end{equation}

where $\mathbf{r}_{i,j}$ denotes the result at the position $(i, j)$ and $\mathbf{x}_{i,j}^{(c)}$ stands for the features at the position $(i, j)$ belonging to channel group $c$. In Fig~\ref{fig:dap} we illustrate the concept of DAP.

We use DAP on the output of our decoder (see Fig~\ref{fig:arch}). In our experiment we set $k=3$. Note that DAP requires the number of feature channels increased by $k\times k$ times, so we increase the output channels of each deconvolution block to $189$ $(21\times 3\times 3)$. For fair comparison, we also evaluate the baseline model with the same number of channels. Results are shown in Table~\ref{tbl:dap}. It is clear that DAP improves the performance by $0.6\%$ while the counterpart model without DAP only obtains marginal gain, which implies DAP may be helpful for feature maps to embed more spatial information.

\begin{table}[h]

\begin{center}
\begin{tabular}{c|c|c}
\hline
Index & Method  & mIoU (\%) \\
\hline
1 & Baseline & 79.0 \\
\hline
2 & Baseline (more channels) & 79.1\\
\hline
3 & DAP (Fig~\ref{fig:dap})  & \textbf{79.6}\\
\hline
\end{tabular}
\end{center}
\caption{Ablation study on the effect of \emph{Densely Adjacent Prediction (DAP)}. The baseline model is in Table~\ref{tbl:ablation} (\#5)}

\label{tbl:dap}
\end{table}

\subsection{Discussions}

\subsubsection{Is Feature Fusion Enhanced?}
At the beginning of Sec~\ref{sec:approach} we demonstrate that feature fusion in our baseline architecture (\emph{GCN} \cite{Peng2017Large}) is ineffective. Only marginal improvements are obtained by fusing low-level features (Level 1 and 2), as shown in Table~\ref{tbl:basicfusion}. We attribute the issue to the semantic and resolution gap between low-level and high-level features. In Sec~\ref{sec:tolowlevel} and Sec~\ref{sec:tohighlevel}, we propose a series of solutions to introduce more semantic information into low-level features and more spatial details into high-level features. 

Despite the improved performance, a question raises: is feature fusion in the framework \emph{really} improved? To justify this, similar to Table~\ref{tbl:basicfusion} we compare several subsets of different feature levels and use them to train original baseline (GCN) and our proposed model (\emph{ExFuse}) respectively. For the ExFuse model, all the 5 approaches in Sec~\ref{sec:tolowlevel} and Sec~\ref{sec:tohighlevel} are used. Table~\ref{tbl:compfusion} shows the results. We find that combined with low-level feature maps (Level 1 and 2) the proposed ExFuse still achieves considerable performance gain ($\sim$1.3\%), while the baseline model cannot benefit from them. The comparison implies our insights and methodology enhance the feature fusion indeed.

Table~\ref{tbl:compfusion} also shows that the proposed model is much better than the baseline in the case that only top-most feature maps (Level 4) are used, which implies the superior high-level feature quality to the original model. Our further study shows that methods in Sec~\ref{sec:tohighlevel} contribute most of the improvement. Empirically we conclude that boosting high-level features not only benefits feature fusion, but also contributes directly to the segmentation performance.

\begin{table}
\begin{center}
\begin{tabular}{c|c|c}
\hline
Feature Levels & Original GCN \cite{Peng2017Large} (\%)& ExFuse (\%) \\
\hline
$\{4\}$ & 73.79& 77.29 \\
\hline
$\{3,4\}$ & 75.97& 78.69 \\
\hline
$\{2,3,4\}$ & 75.98& 79.11\\
\hline
$\{1,2,3,4\}$ & 76.02& 80.04\\
\hline
\end{tabular}
\end{center}

\caption{Comparison of Original GCN \cite{Peng2017Large} and ExFuse on segmentation results using given feature levels. The backbone feature extractor networks are both ResNeXt 101.}

\label{tbl:compfusion}
\end{table}

\subsubsection{Do techniques work in a vanilla U-Net?}
Previously we would like to demonstrate that the proposed perspective and techniques are able to improve one of the state-of-the-art U-Net structure -- GCN. To prove the good generalization of this paper, we apply techniques illustrated above to a vanilla U-Net without GCN module. Performance on PASCAL VOC 2012 is boosted from 72.7 to 79.6 in mIoU. We see that the gap is even bigger (6.9 instead of 4.0), which shows that techniques illustrated above generalize well.

\subsubsection{Could the perspective and techniques generalize to other computer vision tasks?}
Since U-Net structure is widely applied to other vision tasks such as low-level vision \cite{Shen2017Convolutional} and detection \cite{Lin2016Feature}, a question raises naturally: could the proposed perspective and techniques generalize to other tasks? We carefully conducted ablation experiments and observe positive results. We leave detailed discussion for future work.

\section{PASCAL VOC 2012 Experiment}

In the last section we introduce our methodology and evaluate their effectiveness via ablation experiments. In this section we investigate the fully-equipped system and report benchmark results on PASCAL VOC 2012 test set.

To further improve the feature quality, we use deeper ResNeXt 131 as our backbone feature extractor, in which \emph{Squeeze-and-excitation} modules \cite{hu2017squeeze} are also involved. The number of building blocks for Stage 2-5 is $\{8,8,19,8\}$ respectively, which follows the idea of Sec~\ref{sec:lr}. With ResNeXt 131, we get 0.8\% performance gain and achieve 80.8\% mIoU when training with 10582 images from \emph{PASCAL VOC 2012} \cite{Everingham2010The} and \emph{Semantic Boundaries Dataset (SBD)} \cite{Hariharan2011Semantic}, which is 2.3\% better than \emph{DeepLabv3} \cite{Chen2017Rethinking} at the same settings. 

\begin{table}[b]
\begin{center}
\begin{tabular}{|c|ccc|c|}
\hline
Index & ResNeXt 131 & COCO & Flip & mIoU (\%) \\
\hline\hline
1 & (ResNeXt 101) & & & 80.0 \\
\hline
2 & \checkmark & &  & 80.8 \\
\hline
3 & \checkmark & \checkmark  & & 85.4 \\
\hline
4 & \checkmark & \checkmark & \checkmark &  \textbf{85.8} \\
\hline

\end{tabular}
\end{center}
\caption{Strategies and results on PASCAL VOC 2012 validation set }
\label{tbl:improvestrategy}
\end{table}

Following the same procedure as \cite{Chen2016DeepLab,Ghiasi2016Laplacian,Chen2017Rethinking,Islam_2017_CVPR,Peng2017Large,Wang2017Understanding,Zhao2016Pyramid,Lin2016RefineNet,Chen2014Semantic}, we employ Microsoft COCO dataset \cite{Lin2014Microsoft} to pretrain our model. COCO has 80 classes and we only retain images including the same 20 classes in PASCAL VOC 2012 and all other classes are regarded as background. Training process has 3 stages. In stage-1, we mix up all images in COCO, SBD and standard PASCAL VOC 2012 images, resulting in 109892 images for training in total. In stage-2, we utilize SBD and PASCAL VOC 2012 training images. Finally for stage-3, we only employ standard PASCAL VOC 2012 training set. We keep image crop size unchanged during the whole training procedure and all other settings are exactly the same as \cite{Peng2017Large}. COCO pretraining brings about another 4.6\% increase in performance, as shown in Table \ref{tbl:improvestrategy} (\#2 and \#3).

We further average the score map of an image with its horizontal flipped version and eventually get a 85.8\% mIoU on PASCAL VOC 2012 validation set, which is 2.3\% better than DeepLabv3+ \cite{Chen2018Encoder} (Table \ref{tbl:improvestrategy} \#4). 

Resembling \cite{Chen2017Rethinking}, we then freeze the batch normalization parameters and fine tune our model on official PASCAL VOC 2012 \emph{trainval} set. In particular, we duplicate the images that contain hard classes (namely bicycle, chair, dining table, potted plant and sofa). Finally, our ExFuse framework achieves \textbf{87.9\%} mIoU on PASCAL VOC 2012 test set without any DenseCRF~\cite{krahenbuhl2011efficient} post-processing, which surpasses previous state-of-the-art results, as shown in Table \ref{tbl:voctestresults}. For fair comparison, we also evaluate our model using a standard ResNet101 and it achieves 86.2\% mIoU, which is better than DeepLabv3 at the same setting.

\begin{figure*}
\begin{center}
\includegraphics[width=1\linewidth]{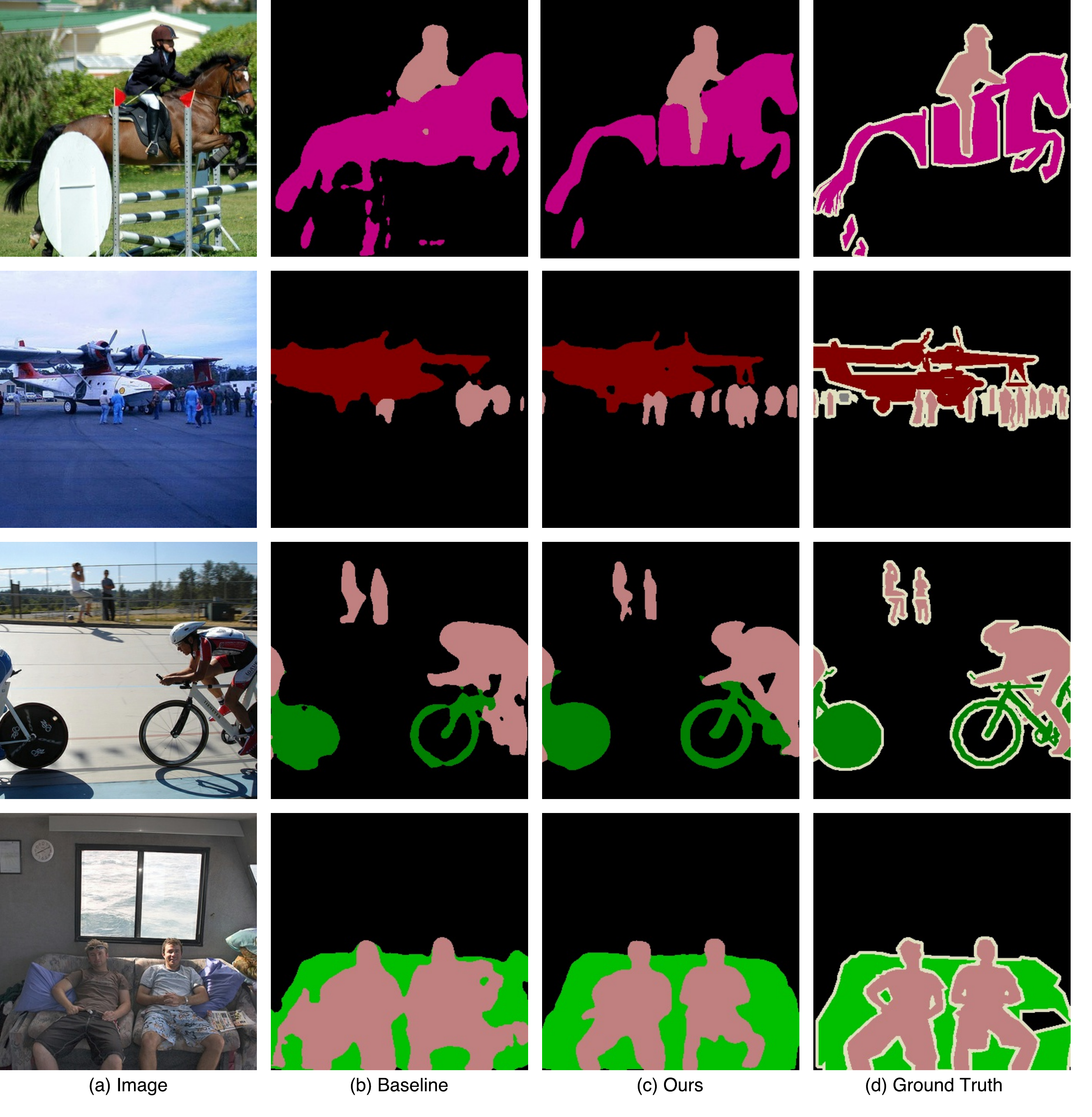}
\end{center}
   \caption{Examples of semantic segmentation results on PASCAL VOC 2012 validation set. (b) is our GCN \cite{Peng2017Large} baseline which achieves 81.0\% mIoU on val set. (c) is our method which achieves 85.4\% on val set, as shown in Table \ref{tbl:improvestrategy} \#3.
}
\label{fig:short}
\end{figure*}

\begin{table}
\begin{center}
\begin{tabular}{|l|c|}
\hline
Method  & mIOU \\
\hline
\hline
Tusimple \cite{Wang2017Understanding}    & 83.1 \\
Large\_Kernel\_Matters \cite{Peng2017Large}    & 83.6 \\
Multipath\_RefineNet \cite{Lin2016RefineNet}   & 84.2 \\
ResNet\_38\_MS\_COCO \cite{Wu2016Wider}  & 84.9 \\
PSPNet \cite{Zhao2016Pyramid}   & 85.4 \\
DeepLabv3 \cite{Chen2017Rethinking}   & 85.7\\
SDN  \cite{Fu2017Stacked} & 86.6 \\
DeepLabv3+ (Xception) \cite{Chen2018Encoder} & 87.8 \\
\hline
\textbf{ExFuse\_ResNet101 (ours)}  & \textbf{86.2} \\
\textbf{ExFuse\_ResNeXt131 (ours)} & \textbf{87.9} \\
\hline
\end{tabular}
\end{center}
\caption{Performance on PASCAL VOC 2012 test set}

\label{tbl:voctestresults}
\end{table}
Fig \ref{fig:short} visualizes some representative results of the GCN \cite{Peng2017Large} baseline and our proposed ExFuse framework. It is clear that the visualization quality of our method is much better than the baseline. For example, the boundary in ExFuse is more precise than GCN.

\section{Conclusions}

In this work, we first point out the ineffective feature fusion problem in current \emph{U-Net} structure. Then, we propose our \emph{ExFuse} framework to tackle this problem via bridging the gap between high-level low-resolution and low-level high-resolution features. Eventually, better feature fusion is demonstrated by the performance boost when fusing with original low-level features and the overall segmentation performance is improved by a large margin. Our \emph{ExFuse} framework also achieves new state-of-the-art performance on PASCAL VOC 2012 benchmark.


\bibliographystyle{splncs}
\bibliography{egbib}
\end{document}